\pdfoutput=1

\documentclass[11pt]{article}

\usepackage[]{ACL2023}

\usepackage{times}
\usepackage{latexsym}

\usepackage{CJKutf8}
\usepackage{microtype}
\usepackage{graphicx}
\usepackage{times}
\usepackage{latexsym}
\usepackage{todonotes}

\usepackage[T1]{fontenc}
\usepackage{multirow}
\usepackage{makecell}
\usepackage{multirow}
\usepackage{booktabs}
\usepackage{colortbl}
\usepackage{tabularx}

\definecolor{lightblue}{RGB}{173,216,230}
\definecolor{skyblue}{RGB}{135,206,235}
\definecolor{cornflowerblue}{RGB}{100,149,237}
\definecolor{steelblue}{RGB}{70,130,180}
\definecolor{royalblue}{RGB}{65,105,225}
\definecolor{navyblue}{RGB}{0,0,128}
\definecolor{green}{HTML}{3C8031}

\usepackage[T1]{fontenc}

\usepackage[utf8]{inputenc}

\usepackage{microtype}

\usepackage{inconsolata}

\title{Discourse Representation Structure Parsing for Chinese}

\author{
Chunliu Wang, Xiao Zhang, Johan Bos\\
CLCG, University of Groningen \\
\texttt{\{chunliu.wang, xiao.zhang, johan.bos\}@rug.nl}
}

\begin{document}
\maketitle
\begin{abstract}
Previous work has predominantly focused on monolingual English semantic parsing. We, instead, explore the feasibility of Chinese semantic parsing in the absence of labeled data for Chinese meaning representations.
We describe the pipeline of automatically collecting the linearized Chinese meaning representation data for sequential-to-sequential neural networks.
We further propose a test suite designed explicitly for Chinese semantic parsing, which provides fine-grained evaluation for parsing performance, where we aim to study Chinese parsing difficulties.
Our experimental results show that the difficulty of Chinese semantic parsing is mainly caused by adverbs. Realizing Chinese parsing through machine translation and an English parser yields slightly lower performance than training a model directly on Chinese data.

\end{abstract}

\section{Introduction}

Semantic parsing is the task of transducing natural language text into semantic representations, which are expressed in logical forms underlying various grammar formalisms, such as abstract meaning representations (AMR, ~\citealt{wang-etal-2020-amr, bevilacqua-etal-2021-one}), minimal recursion semantics (MRS, ~\citealt{horvat-etal-2015-mrs}), and Discourse Representation Theory (DRT, ~\citealt{Kamp1993}).
In this work, we explore the feasibility of parsing Chinese text to semantic representation based on Discourse Representation Structures (DRSs, ~\citealt{bos15-boxer, van-2018-exploring}), which are meaning representations proposed from DRT, a recursive first-order logic representation comprising of discourse referents (the entities introduced in the discourse) and relations between them. 

Several neural parsers for DRS have been recently developed \citep{fancellu-2019-semantic, evang-2019-transition,van-2019-linguistic, liu-2019-discourse-representation, wang-2020, van-2020-character} and reached remarkable performance, but mostly focused on monolingual English or some language using the Latin alphabet. Meaning representations are considered to be language-neutral, and texts with the same semantics but in different languages have the same meaning representation. The literature presents several examples of parsing multilingual text by training on monolingual English semantic representations \citep{ribeiro-etal-2021-smelting}.

For the reason of relatively limited amounts of labeled gold-standard multilingual meaning representation data, multilingual text parsing relies on the source of silver English meaning representation data.
As long as the meanings are expressed in a language-neutral way, this is a valid approach. However, named entities aren't usually, because they can (a) have different orthography for different languages using the same alphabet (in particular for location names, e.g., Berlin, Berlijn, Berlino, Berlynas) or (b) be written with a completely different character set, as is the case for Chinese. 

\begin{figure}[htbp]
\centering
\includegraphics[scale=.70]{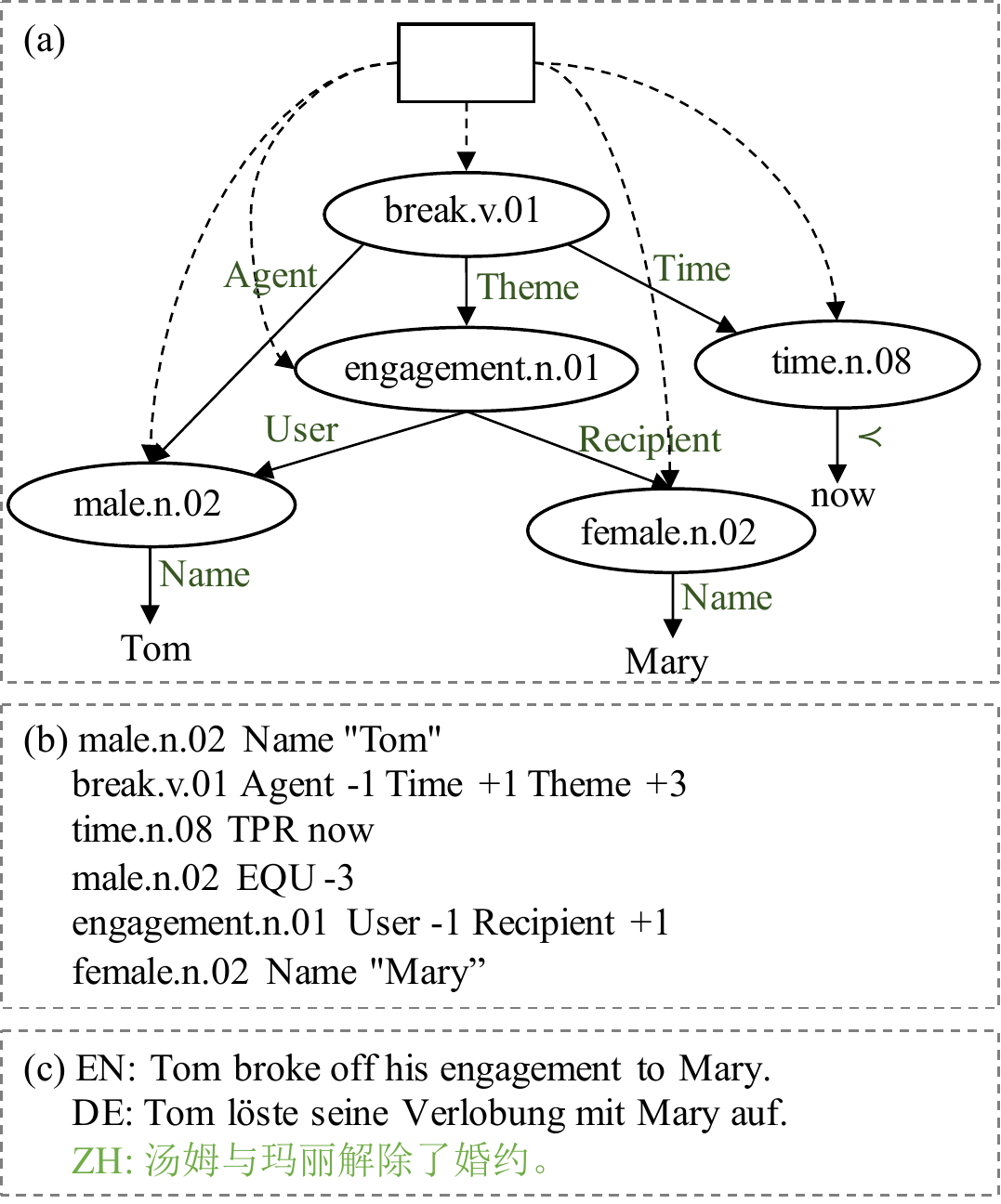}
\caption{DRS in (a) graph format, (b) sequential box notation  and (c) corresponding multilingual texts for English, German and Chinese.} 
\label{fig:drs}
\end{figure}

Figure~\ref{fig:drs} shows a (nearly) language-neutral meaning representation for a simple English sentence.
For non-English Latin alphabet languages, the named entities in the text are usually consistent with English, and the meaning in the form of a graph structure of the corresponding Discourse Representation (Discourse Representation Graph, DRG) would be identical to these languages \citep{bos2021variable}, as shown in Figure~\ref{fig:drs}. However, it would be rather absurd to expect a semantic parser for Chinese to produce meaning representations (with interlingual WordNet synsets) where proper names are anchored using the Latin alphabet using English (or any other language for that matter) orthography. We need to keep this important aspect in mind when evaluating semantic parsers for languages other than English.

However, for non-Latin alphabet languages, such as the widely used language of Chinese, is it feasible to use English meaning representation as the meaning representation of Chinese?
Our objective is to investigate whether Chinese semantic parsing can achieve the same performance as English semantic parsing while using the same amount of data.
We try to investigate whether it is necessary to develop a dedicated parser for Chinese, or whether it is possible to achieve a similar performance using an English parser by leveraging machine translation (MT) on Chinese.
We provide inexpensively acquired silver-standard Chinese DRS data to implement our exploration: 
(1) We collect Chinese and English aligned texts from the Parallel Meaning Bank (PMB, \citealt{abzianidze-parallel}), which provides parallel multilingual corpora including corresponding English meaning representation expressed in DRSs.
(2) We leverage GIZA++ \citep{giza-2003-systematic} to align the word-segmented Chinese and English to obtain Chinese-English named entity alignment pairs, the resulting named entities are used to replace the named entities in our English semantic representation.
(3) We train two monolingual parsers on the two languages separately, and then provide a set of fine-grained evaluation metrics to make better comparison between parsers.
We aim to answer the following questions: 
\begin{enumerate}

     \item Can existing DRS parsing models  
           achieve good results for Chinese? (RQ1)
     
     \item What are the difficulties in semantic parsing for Chinese? (RQ2)
    
     \item Is it feasible to use machine translation and an English parser to parse Chinese? How is it different from designing a special parser for Chinese? (RQ3)

     \item How to conduct more fine-grained evaluation of experimental results and reduce the workload of manual evaluation? (RQ4)
         
\end{enumerate}

\section{Background}
\subsection{Discourse Representation Structure} \label{drs_intro}

\begin{figure}[htbp]
\centering
\includegraphics[scale=.65]{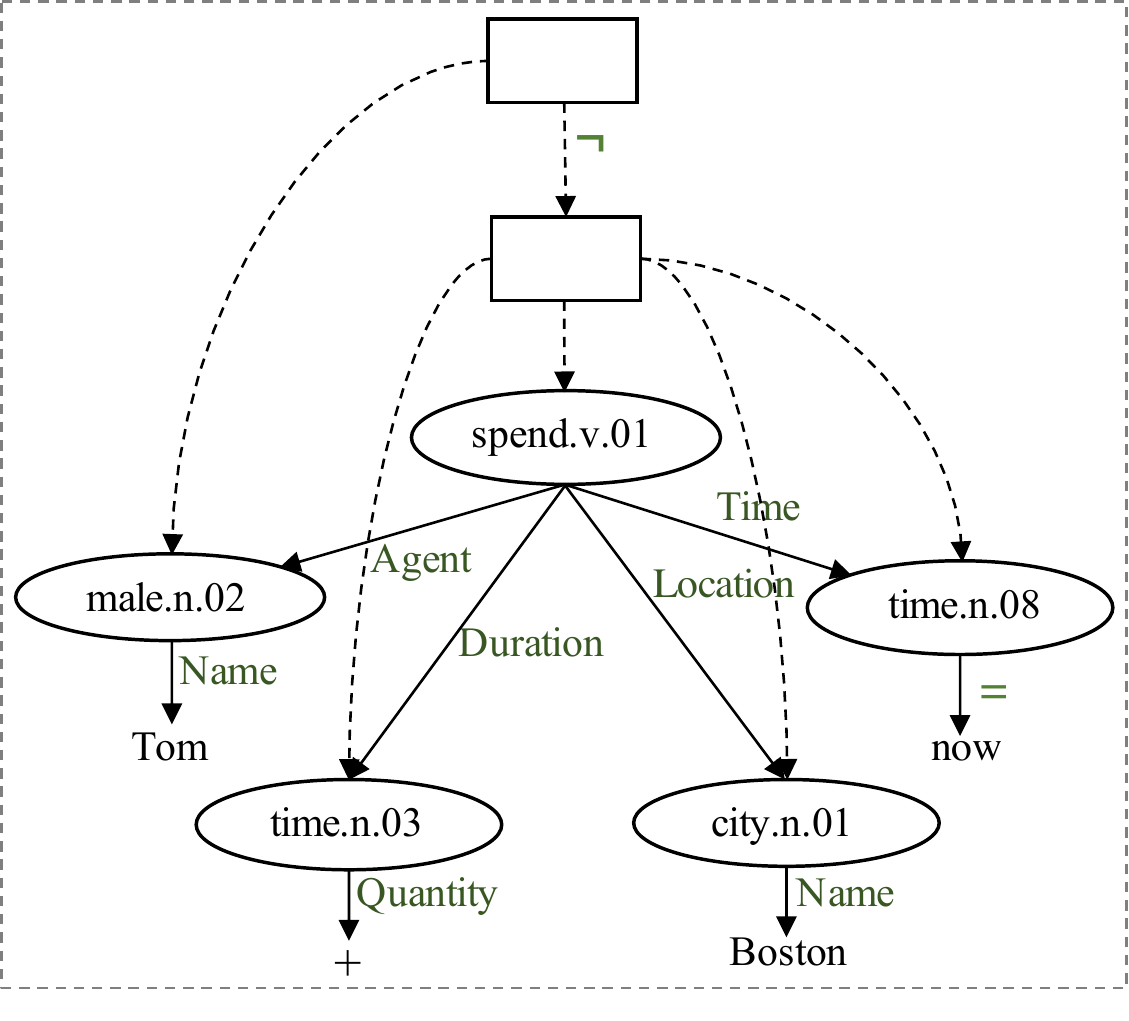}
\caption{An example of a DRG with negation for sentence: "Tom doesn't spend much time in Boston."} 
\label{fig:sbn-discourse}
\end{figure}

DRS, as a kind of formal meaning representation, can be used to represent the semantic meaning of sentences and discourse. For the wide coverage of linguistic phenomena at quantification, negation, reference resolution, comparatives, discourse relations, and presupposition, DRT and DRS possess stronger semantic representation power than AMR. 
A DRS comprises discourse referents and conditions. However, some variants of DRS formats have been introduced in recent years, the format we employ throughout our work being one of them. We use a simplified DRS, which can be called Discourse Representation Graph (DRG) or Simplified Box Notation (SBN; ~\citealt{bos2021variable}). It discards explicit discourse references and variables while maintaining the same expressive power, as shown in Figure~\ref{fig:sbn-discourse}. 

As introduced by \citet{bos2021variable}, DRS allows two kinds of representations: graph and sequential notation (Figure \ref{fig:drs}). There are five types of semantic information involved in DRS: concepts (\texttt{read.v.01}, \texttt{paper.n.02}, \texttt{new.a.01}, \texttt{...}), roles (\texttt{Agent}, \texttt{Theme}, \texttt{Time}, \texttt{...}), constants (\texttt{speaker}, \texttt{hearer}, \texttt{now}, \texttt{...}), comparison operators ($=$, $\prec$, $\sim$, \texttt{...}) and discourse relations (\texttt{NEGATION}, \texttt{CONTINUATION}, \texttt{CONTRAST}, \texttt{...}), where concepts and roles are represented by WordNet synsets \cite{Fellbaum2000WordNetA} and VerbNet thematic relations \cite{kipper-etal-2006-extending} respectively.

\begin{figure*}[t]
\centering
\includegraphics[scale=.70]{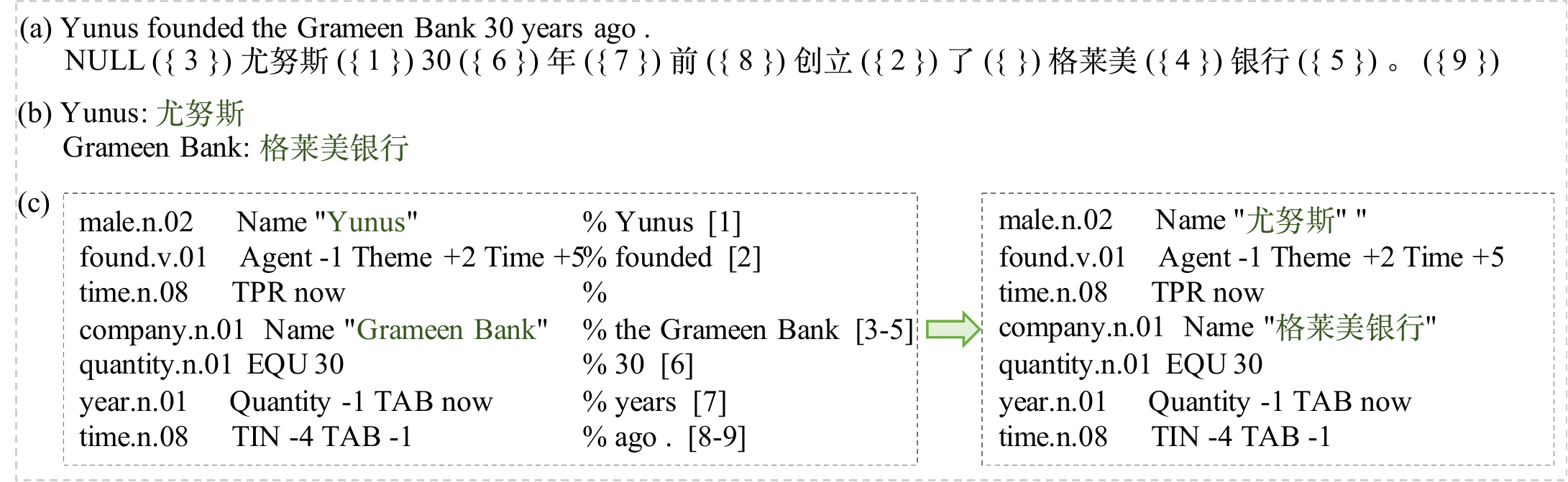}
\caption{(a) Alignment tokens obtained by GIZA++ tool for English  (\textit{"Yunus founded the Grameen Bank 30 years ago."}) and Chinese ("\begin{CJK}{UTF8}{gbsn}尤努斯30年前创立了格莱美银行。\end{CJK}"), (b) aligned named entities dictionary in above texts, (c) same meaning representations with different named entities for English text and Chinese text respectively.} 
\label{fig:en-zh-drs}
\end{figure*}

\subsection{DRS parsing}

DRS parsing was originally applied to English and has been continuously extended to other Latin languages. Initially, rule-based systems were predominantly utilized by early parsers for analyzing small English texts \citep{Johanson1986parsing, ASHER1988parsing, Bos2004ComputationalSI, bos-2008-wide, Bos2015OpenDomainSP}. The first version of GMB \cite{BasileBosEvangVenhuizen2012EACL} which provides English texts with DRS, is built on Boxer \cite{bos-2008-wide}.  
With the release of PMB \cite{abzianidze-parallel} and the propose of the first shared tasks \cite{abzianidze-etal-2019-first}, related research keeps growing, with a focus on deep learning models \cite{evang-2019-transition, fancellu-2019-semantic, van-2018-exploring, van-2020-character, liu-2019-discourse-representation}. 
The target languages have also expanded to other languages: German, Italian, Dutch and Chinese \cite{shen-evang-2022-drs, poelman-etal-2022-transparent, wang-2020, liu2021Univeral}. 
Translation has been utilized in two manners when dealing with cross-lingual parsing: the first involves translating other languages into English and then employing an English parser, while the second involves translating English into other languages and training a parser specific to that language \cite{liu2021Univeral}.
In this paper, we use the existing Chinese-English parallel corpus to design a specific parser for Chinese, and compare the performance of the parser with the first method.

\section{Data Creation \label{data}}

In previous work, for non-English parsing tasks, the semantic representation of English is usually directly used as the semantic representation of the target language, but most of these works focus on Latin languages~\citep{fancellu-2019-semantic, ribeiro-etal-2021-smelting}.
For non-Latin languages such as Chinese, named entities are not language-neutral, as illustrated in the work of \citet{wang-2020}, and are quite different from named entities in English texts.
To design a more reasonable Chinese parser, we first focus on replacing the named entities in the English semantic representation with Chinese, so that the parser can parse out the Chinese named entities corresponding to the text content according to different texts.

To achieve our goal, we use the data of PMB, the largest parallel corpus of DRS data available, as our experimental object. From the PMB, English-Chinese parallel texts and DRS data for English texts are collected. Based on that, we propose a pipeline to obtain Chinese DRS for Chinese text. Our pipeline has three steps: 
(1) using tokenizers tools to segment Chinese and English text data;
(2) utilizing the English-Chinese alignment tool to obtain the alignment tokens between Chinese and English texts;
(3) replacing named entities in English DRS with Chinese named entities.
Figure~\ref{fig:en-zh-drs} shows our processing pipeline.

\subsection{Text Tokenizers}
Preprocessing data with a tokenizer is an important step in the pipeline because the alignment of Chinese and English texts needs to act on the data after tokenization. At the same time, since the quality of upstream results directly affects downstream performance, the quality of text segmentation also directly affects the correctness of Chinese and English text alignment.
In this work, we use Moses~\citep{koehn-etal-2007-moses} for English, which is advanced and widely used. It is a collection of complex normalization and segmentation logic that works very well for structured languages like English.
For Chinese, we choose HanLP~\citep{he-choi-2021-stem}, which is an efficient, user-friendly and extendable tokenizer. Different from a widely used Jieba tokenizer, HanLP is based on the CRF algorithm.
It takes into account word frequency and context at the same time, and can better identify ambiguous words and unregistered words.

\subsection{English-Chinese Alignment} 

In order to realize the replacement of named entities in English semantic representation with Chinese named entities, it is very important to obtain the correct alignment of Chinese and English texts, especially the alignment of named entities in the two texts.
In order to quickly and effectively obtain the alignment data in Chinese and English, we choose the GIZA++ word aligning tool.
GIZA++ is the most popular statistical alignment and MT toolkit~\citep{10.3115/1075218.1075274}, which implements the lexical translation models of \citet{brown-etal-1993-mathematics} (IBM Models), and the Hidden-Markov alignment Model~\citep{vogel-etal-1996-hmm}, trained using expectation-maximization (EM). 
GIZA++ is highly effective at aligning frequent words in a corpus, but error-prone for infrequent words.

\subsection{Replacing Named Entities}

The last step to obtain the Chinese semantic representation is to replace the named entities in the English DRS with Chinese named entities. First, the English named entities in DRS data can be easily obtained according to the edge types between two nodes. When the edge type is \texttt{Name}, the output nodes are named entities in the DRG. 
After processing the Chinese and English texts with the GIZA++ tool in the second step, we can obtain alignment tokens between Chinese and English. On this basis, a named entity alignment dictionary can be obtained, and then the English named entities in the DRS data can be replaced with Chinese named entities based on this dictionary.

\section{Methodology}

\subsection{Neural Models\label{my_model}}

We adopt Recurrent Neural Networks (RNN) equipped with Long Short-Term Memory units (LSTM; ~\citealt{lstm1997}) as our baseline models. 
Following the work of \citet{van-noord-etal-2020-character}, 
we use frozen mBERT \citep{devlin-etal-2019-bert} embeddings to initialize the encoder.
An attention-based LSTM architecture is used for the decoder, where the attention memory is the concatenation of the attention vectors among all the input tokens.
In addition, the copy mechanism~\citep{gu-etal-2016-incorporating, gulcehre-etal-2016-pointing} is added to the decoder, which can integrate the attention distribution into the final vocabulary distribution. 
The copy mechanism favors copying tokens from the source text into the target text instead of generating all target tokens only from the target vocabulary.

\subsection{Evaluation}

\begin{figure*}[th]
\centering
    \includegraphics[scale=.82]{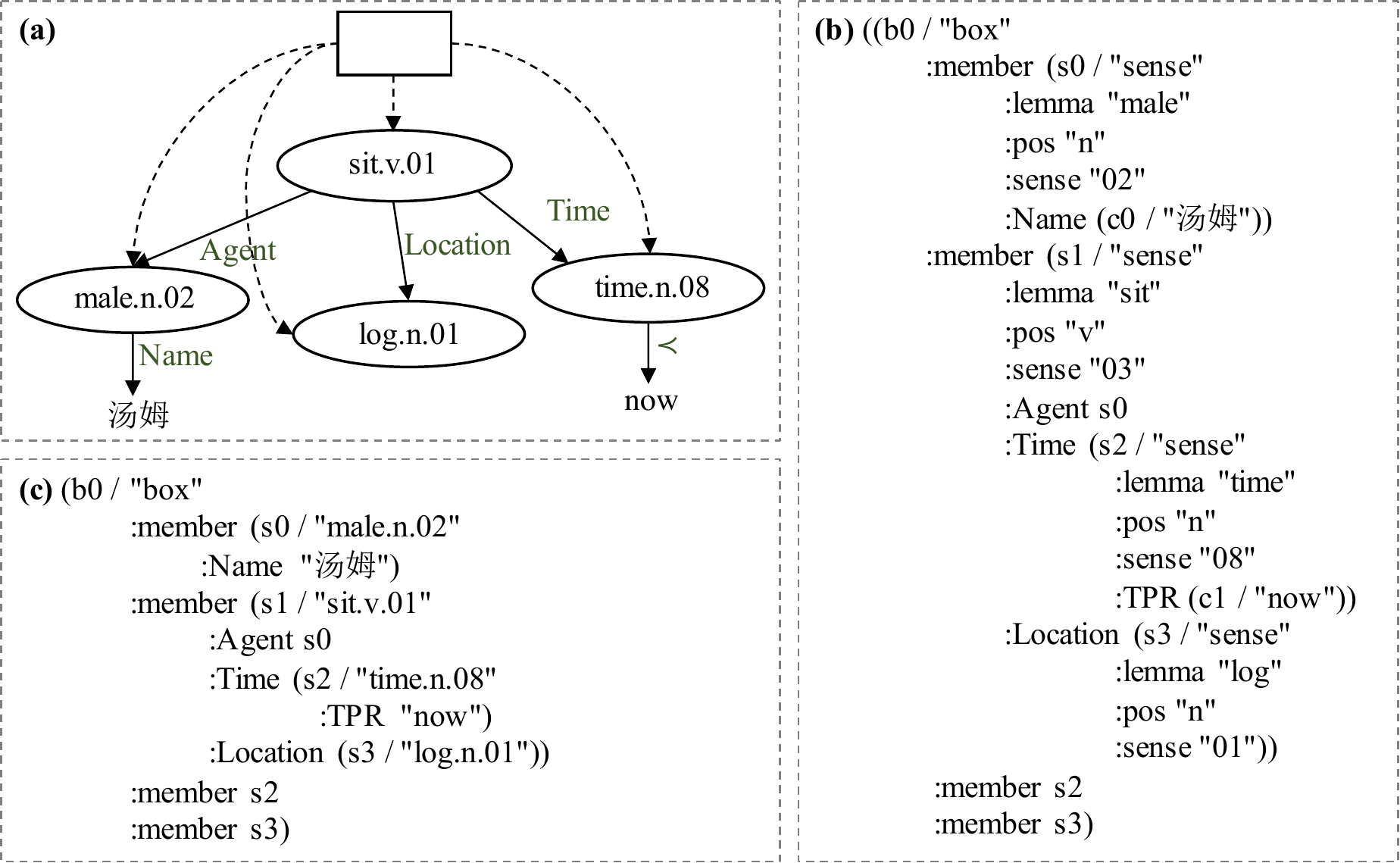}
\caption{(a) Graph structured DRS for Chinese sentence:"~\begin{CJK}{UTF8}{gbsn}汤姆坐在一根圆木上。\end{CJK}". (b) Penman format of DRG with fine-grained WordNet synsets used for evaluation~\citep{wessel2022}. (c) Penman format of DRG with coarse-grained WordNet synsets used for evaluation (Ours).}
\label{fig:penman}
\end{figure*}

Given a document to the DRS parser, it will generate variable-free sequential notation DRS as shown in Figure~\ref{fig:drs}(b). 
The evaluation tool for DRS parsing task was recently proposed by \citet{wessel2022} and is based on the AMR standard evaluation tool Smatch~\citep{cai-knight-2013-smatch}.
By converting a sequential DRS into DRG, Penman notation format data~\citep{kasper-1989-flexible} can be obtained, as shown in Figure~\ref{fig:penman}~(b), and then Smatch can be used to compute F-scores based on matching triples between system output and gold meanings.

However, we note that the scores given by the above evaluation tool have two flaws: 
(1) the evaluation scores are too inflated, and it is difficult to detect the differences between different parsers.
(2) the evaluation tool only gives an overall score without evaluating the different types of constituent elements in the DRS, it is difficult to quantitatively determine what is the difficulty of the parser in the parsing process.
Based on that, we propose to compress evaluation scores to improve the above evaluation methods and further propose fine-grained evaluation metrics for different subtasks according to different types of components in DRS.

\subsubsection{Overall Evaluation}

Our improvement strategy is mainly aimed at the representation of the Penman format of DRG. We mainly improve on two points, one is WordNet synsets representation, and the other is constants representation.

In the previous evaluation method, the WordNet synsets in Penman format are fine-grained during the evaluation process, and the WordNet synsets are divided into three parts (lemma, pos, number) according to their constituents. 
On this basis, even if the parser generates wrong concepts, such as \texttt{time.n.08} and \texttt{time.n.01}, the Smatch still obtains a similar inflated F1 score.
To this end, we change the WordNet synsets in the Penman format to a coarse-grained representation to strictly evaluate WordNet synsets qualities generated by parsers, as shown in Figure~\ref{fig:penman} (c).
In addition, we have also modified the constant representation in Penmen format, such as the constant \texttt{now} shown in the figure, because the variable \textit{c} is added to the constant, making the triples in Penman format redundant, which also makes the F1-score higher to a certain extent. By omitting the \textit{c} variable as shown in Figure~\ref{fig:penman} (c), we further compress the F1-score.

\subsubsection{Fine-grained Evaluation}

To evaluate the quality of specific subtasks in DRS parsing, we imitate the fine-grained metrics for AMR parsing task~\citep{damonte-etal-2017-incremental, zhang-etal-2019-amr} to DRS parsing. In order to make them compatible with DRS, we make some changes based on the data characteristics of DRS. 
Our fine-grained metrics consist of three categories in total: \emph{graph-level}, \emph{node-level} and \emph{edge-level}. Each category includes more fine-grained evaluation metrics.
All the metrics are proposed based on the semantic information types involved in DRS (see Section~\ref{drs_intro}).

In \textbf{\emph{graph-level}} evaluation, \texttt{No Roles}, \texttt{No Discourse}, \texttt{No Operators} and \texttt{No Senses} are used to represent the Smatch scores of the DRG in Penman format ignoring Roles, Discourse, Operators and Senses respectively. In theory, they are Smatch's coarse-grained scores, which are higher than the original Smatch scores.

\begin{table*}[t]
\centering
\scriptsize
\resizebox{\textwidth}{!}{
\begin{tabular}{ccp{8cm}}
\toprule
\textbf{Alignment Error}  & \textbf{Data Type} & \multicolumn{1}{c}{\textbf{Example}}\\ 
\midrule
\multirow{8}*{Dislocation} & Chinese & \begin{CJK}{UTF8}{gbsn}\textcolor{skyblue}{梅尔·卡玛津} 是 \textcolor{blue}{天狼星} 的 \textcolor{cornflowerblue}{执行官}。\end{CJK}  \\
& English & \textcolor{skyblue}{Mel Karmazin} is an \textcolor{cornflowerblue}{executive} of \textcolor{blue}{Sirius} .\\
& Wrong & male.n.02 Name "\begin{CJK}{UTF8}{gbsn}梅尔·卡玛津\end{CJK}" be.v.08 Theme -1 Time +1 Co-Theme +2 time.n.08 EQU now person.n.01 Role +1 executive.n.01 Of +1 company.n.01 Name "\begin{CJK}{UTF8}{gbsn}\textcolor{red}{执行官}\end{CJK}"\\
& Corrected & male.n.02 Name "\begin{CJK}{UTF8}{gbsn}梅尔·卡玛津\end{CJK}" be.v.08 Theme -1 Time +1 Co-Theme +2 time.n.08 EQU now person.n.01 Role +1 executive.n.01 Of +1 company.n.01 Name "\begin{CJK}{UTF8}{gbsn}\textcolor{green}{天狼星}\end{CJK}"\\
\hline
\multirow{6}*{Character Exclusion} & Chinese & \begin{CJK}{UTF8}{gbsn}什么 乐队 唱 了 “ \textcolor{blue}{快乐 在 一起} ” 这 首 歌 ？\end{CJK} \\
& English & What group sang the song " \textcolor{blue}{Happy Together} " ?\\
& Wrong   & group.n.01 Name ? sing.v.02 Agent -1 Time +1 Theme +2 time.n.08 TPR now song.n.01 EQU +1 music.n.01 Name "\begin{CJK}{UTF8}{gbsn}\textcolor{red}{快乐一起}\end{CJK}"\\
& Corrected &  group.n.01 Name ? sing.v.02 Agent -1 Time +1 Theme +2 time.n.08 TPR now song.n.01 EQU +1 music.n.01 Name "\begin{CJK}{UTF8}{gbsn}\textcolor{green}{快乐在一起}\end{CJK}"  \\
\hline
\multirow{7}*{Character Inclusion} & Chinese &  \begin{CJK}{UTF8}{gbsn}\textcolor{blue}{卢瑟福·海斯} \textcolor{skyblue}{1822} 年 出生 于\textcolor{cornflowerblue}{俄亥俄州} 。\end{CJK} \\
& English & \textcolor{blue}{Rutherford Hayes} was born in \textcolor{cornflowerblue}{Ohio} in \textcolor{skyblue}{1822} .\\
& Wrong & male.n.02 Name "\begin{CJK}{UTF8}{gbsn}\textcolor{red}{卢瑟福·海斯1822}\end{CJK}" time.n.08 TPR now bear.v.02 Patient -2 Location +1 Time +2 state.n.01 Name "\begin{CJK}{UTF8}{gbsn}俄亥俄州\end{CJK}" time.n.08 YearOfCentury 1822 TIN -3\\
& Corrected &  male.n.02 Name "\begin{CJK}{UTF8}{gbsn}\textcolor{green}{卢瑟福·海斯}\end{CJK}" time.n.08 TPR now bear.v.02 Patient -2 Location +1 Time +2 state.n.01 Name "\begin{CJK}{UTF8}{gbsn}俄亥俄州\end{CJK}" time.n.08 YearOfCentury 1822 TIN -3 \\
\hline
\multirow{7}*{Nationality} &  Chinese & \begin{CJK}{UTF8}{gbsn}我 不 是 \textcolor{blue}{爱尔兰人} 。\end{CJK}\\
& English & I am not \textcolor{blue}{Irish} .\\
& Wrong & person.n.01 EQU speaker NEGATION \textless{}1 time.n.08 EQU now be.v.03 Theme -2 Time -1 Source +1 country.n.02 Name \textcolor{red}{""}\\
& Corrected & person.n.01 EQU speaker NEGATION \textless{}1 time.n.08 EQU now be.v.03 Theme -2 Time -1 Source +1 country.n.02 Name \textcolor{green}{"ireland"} \\ 
\bottomrule
\end{tabular}}
\caption{Alignment errors illustrated by four examples. In Chinese and English texts, words of the same color indicate correct alignment between them. Inbformation marked in red is the wrong named entity obtained by the GIZA++ tool. Text in green indicates the correct named entity in the corrected DRS.}
\label{table: Alignment error}
\end{table*}

In \textbf{\emph{node-level}} evaluation, we compute F-score on the list of parsed information types (such as roles, constants, and discourse relations) instead of using Smatch. 
Note that different from the metrics in the AMR parsing task, concepts in DRS are represented by WordNet synsets, so \texttt{Concepts} can be evaluated more finely by part-of-speech (\texttt{noun}, \texttt{adjective}, \texttt{adverb} and \texttt{verb}). 
\texttt{Discourse} detects all discourse relation labels except \textit{NEGATION} since it is more common and specific in DRS than other discourse relations labels, the \texttt{Negation} metric is used for evaluation to detect \textit{NEGATION} edge label alone.
In addition, \texttt{Member} metric is added to evaluate the ratio of the generated concepts. In DRG, \textit{member} represents the edge label connecting the \textit{BOX} node and the concepts node, i.e., the dashed line as shown in Figure~\ref{fig:penman} (a).

For \textbf{\emph{edge-level}} evaluation, we focus on calculating the F-score based on the number of matching triples in the parsed DRG and the gold DRG.
For example, \texttt{Names} in edge-level is a metric that considers the relations between concepts nodes and named entities, which differs from the metric of \texttt{Names} in node-level, which only considers the concepts labeled with \textit{Name} and ignores the accuracy of named entities themselves. ~\footnote{Our evaluation suite is available at: \url{https://github.com/wangchunliu/SBN-evaluation-tool}.} 

\section{Experiments}

\subsection{Dataset}

\begin{table*}[h]
\centering
\resizebox{\textwidth}{!}{
\begin{tabular}{ccl}
\toprule
\textbf{Alignment Error} & \textbf{Reason}  & \multicolumn{1}{c}{\textbf{Example}}\\ 
\midrule
\multirow{4}*{Named-entities} & \multirow{2}*{Jieba} & 
English: Melanie killed a spider with her hand . \\
& & Chinese: \begin{CJK}{UTF8}{gbsn}\textcolor{red}{媚兰用} ({ 1 5 6 }) 手 ({ 7 }) 杀死 ({ 2 })  了 ({ })  一只 ({ 3 })  蜘蛛 ({ 4 }) 。 ({ 8 }) \end{CJK}\\
\cline{2-3}
&  \multirow{2}*{HanLP} & English: Melanie killed a spider with her hand . \\
& & Chinese: \begin{CJK}{UTF8}{gbsn}\textcolor{green}{媚兰} ({ 1 6 }) 用 ({ 5 }) 手 ({ 7 }) 杀死 ({ 2 }) 了 ({ }) 一 ({ 3 }) 只 ({ }) 蜘蛛 ({ 4 }) 。 ({ 8 }) \end{CJK} \\
\midrule
\multirow{4}*{Information units} & \multirow{2}*{gold data} & The ground floor was flooded . \\
& & Chinese: \begin{CJK}{UTF8}{gbsn}\textcolor{red}{一楼 ({ 1 })} 被 ({ }) \textcolor{red}{淹 ({ 2 3 4 5 })} 了 ({ }) 。 ({ 6 }) \end{CJK}\\
\cline{2-3}
& \multirow{2}*{all data} & English: The ground floor was flooded . \\ 
& & Chinese: \begin{CJK}{UTF8}{gbsn}\textcolor{green}{一楼 ({ 1 2 3 })} 被 ({ 4 }) 淹 ({ 5 }) 了 ({ }) 。 ({ 6 }) \end{CJK}\\
\bottomrule
\end{tabular}}
\caption{Impact of different tokenizers and data sizes on GIZA++ performance.}
\label{table: tokenizer}
\end{table*}

We collect all Chinese-English text pairs in the PMB.
According to the quality label of English DRS, we divide the data into gold data and silver data, and randomly split the test set and development set from the gold data. Since PMB data may contain duplicate data, before splitting, we first filter the duplicate data.
Then we merge the remaining gold data and silver data as our training set, and get a total of 137,781 training instances, 
1,000 development instances and 1,000 test instances, each instance contains English DRS data, corresponding English text, and Chinese text.~\footnote{Our data and code are available at: \url{https://github.com/wangchunliu/Chinese-SBN-parsing}.} 

After splitting the data, we use the pipeline introduced in Section~\ref{data} to process our Chinese and English texts to get the Chinese and English word alignment data, and then replace the named entities in the English DRS with Chinese. However, we noticed that not all replacements were successful. We classified the wrong replacement types into four types, as shown in Table~\ref{table: Alignment error}. 
These errors are mainly caused by GIZA++ alignment errors when aligning Chinese and English text words.
Among them, the fourth type of error is quite special. In our experiment, we directly ignore the location named entities used to refer to nationality and do not replace them with Chinese named entites.
In order to reduce the work of manual correction and make the work reproducible, We only fix incorrect named entity replacements in the test set, where 26 of the 1000 test set instances require manual correction of named entities.

\subsection{Settings}

For tokenizers, we use Moses~\citep{koehn-etal-2007-moses} and HanLP~\citep{he-choi-2021-stem} on English and Chinese respectively.
We observe that the HanLP tokenizer outperforms Jieba\footnote{https://github.com/fxsjy/jieba}, a tokenizer widely used in Chinese, in segmenting text containing named entities.
This is an important indicator for selecting a tokenizer, because getting the correct Chinese and English named entity pairs is our main goal.
In addition, we observed that HanLP's segmentation results also outperformed Jieba's tokenizer on text containing traditional Chinese characters, while the Chinese data in PMB contains traditional Chinese characters. This is also one of the reasons for choosing the HanLP tokenizer.
At the top of Table~\ref{table: tokenizer}, we show the difference in name entities between the Jieba tokenizer and the HanLP tokenizer.
In addition, we give an example of the impact of different sizes of training data on the alignment performance of GIZA++ at the bottom of Table~\ref{table: tokenizer}, and the results show that it is almost impossible to achieve correct alignment using only gold data.

\begin{table}[h]
\centering
\setlength{\tabcolsep}{4pt}
\resizebox{\columnwidth}{!}{
\begin{tabular}{lccccc}
\toprule
& \multicolumn{3}{c}{\textbf{Document-level}} & \multicolumn{2}{c}{\textbf{Word-level}} \\
\hline
\textbf{Data} & \textbf{Train}  & \textbf{dev} & \textbf{test} & \textbf{src} & \textbf{tgt} \\ 
\midrule
\textit{English} & 137,781 & 1,000 & 1,000 & 38,441 & 39,761\\
\textit{Chinese} & 137,781 & 1,000 & 1,000 & 42,446 & 41,734\\
\bottomrule
\end{tabular}}
\caption{Document statistics and vocabulary sizes.}
\label{table:vocab}
\end{table}

All experiments are implemented based on OpenNMT \citep{klein-etal-2017-opennmt}. 
For the vocabulary, we construct vocabularies from all words, the vocabulary sizes as shown in Table~\ref{table:vocab}. 
The hyperparameters are set based on performance on the development set. We use SGD optimizer with the initial learning rate set to 1 and decay 0.8. 
In addition, we set the dropout to 0.5 at the decoder layer to avoid overfitting with batch size 32.

\begin{table}[t]
\centering
\setlength{\tabcolsep}{4pt}
\begin{tabular}{lccc}
\toprule
\textbf{Metric}  &\textbf{EN}  & \textbf{ZH} & \textbf{ZH$\to$EN$_{zh}$}  \\
\midrule
Smatch$_1$  & 91.0 & 86.0 & 84.7 \\
Smatch$_2$  & 88.9 & 83.8 & 81.7 \\
\midrule
Well-formed & 99.8 & 99.7 & 99.7 \\
\hline
\textbf{Graph-level} \\
No Roles     & 90.0 & 85.5 & 84.2 \\
No Discourse& 89.5 & 83.9 & 82.7 \\
No Operators & 89.5 & 84.7 & 83.4 \\
No Senses    & 91.9 & 85.6 & 84.7 \\
\bottomrule
\end{tabular}
\caption{F-scores with Smatch on the test set of semantic parsers. Note: Smatch$_1$ and Smatch$_2$ represent the original evaluation~\citep{wessel2022} and our improved evaluation.
}
\label{table:overall_results}
\end{table}

\subsection{Main Results}

Table~\ref{table:overall_results} shows the results obtained by the parsers with Smatch, which gives the overall performance for different parsers.
The first parser (\textbf{EN}) is trained on the English dataset based on the model introduced in Section~\ref{my_model}. The Smatch$_1$ result of our English parser is slightly lower than the results of \citet{wessel2022}, which we believe is due to slightly different training, development and test set instances.
The result of Smatch$_2$ is significantly lower than the result of Smatch$_1$, indicating that the F1-score has been significantly compressed and will not be too inflated (see Section~4).

The Chinese parser~(\textbf{ZH}) is trained on the data created by the pipeline introduced in Section~\ref{data}. The results show that the performance of the Chinese parser is lower than the English parser in all overall evaluation metrics. 
\textbf{ZH$\to$EN$_{zh}$} shows the performance by using the English parser on English text translated from Chinese text instead of training a dedicated model for Chinese text. 
The only unreasonable point is that the model will generate English named entities, which may not be recognized as the correct Chinese semantic representation.

The smatch$_1$ scores and the smatch$_2$ scores show that the Chinese parser outperforms using the ZH$\to$EN$_{zh}$ approach. 
For the metrics \texttt{No Senses} and \texttt{No Roles}, the evaluation results have been significantly improved compared with Smatch$_2$. 
This shows that Concepts and Roles have a greater impact on evaluation results than Discourse and Operators.
It is worth noting that the performance difference between the Chinese and English parsers is about five percentage points across all metrics, while the difference between the ZH and the ZH$\to$EN$_{zh}$  narrows at the graph-level metrics compared to Smatch$_2$ score.

\subsection{Fine-grained Results and Analysis}

To further explore the performance of parsers, we apply our proposed fine-grained evaluation metrics to the results of two parsers. Tabel~\ref{table:fine_results} shows the fine-grained evaluation performance of different component types based on DRG at node-level and edge-level.

\begin{table}[t]
\centering
\setlength{\tabcolsep}{4pt}
\begin{tabular}{rlccc}
\toprule
&\textbf{Metric}  &\textbf{EN}  & \textbf{ZH} & \textbf{ZH$\to$EN$_{zh}$}  \\
\midrule
\textbf{Node}  & Names & 70.8 & \underline{66.0} & 67.7 \\
&Negation  & 92.3 & 88.7 & 88.8 \\
&Discourse & 86.0 & 80.4 & \underline{75.2} \\
&Roles     & 89.2 & 84.0 & 84.9 \\
&Members   & 97.5 & 95.4 & 95.9 \\
&Concepts  & 81.2 & \underline{73.3} & 74.4 \\
&\makecell[r]{\textit{noun}} & 87.1 & \underline{82.1} & 83.3 \\
&\makecell[r]{\textit{adj}}   & 73.3 & 54.2 & \underline{52.5} \\
&\makecell[r]{\textit{adv}}   & 76.8 & \underline{35.3} & 45.5 \\
&\makecell[r]{\textit{verb}} & 59.7 & \underline{45.5} & 47.2 \\
\hline
\textbf{Edge} &Roles      & 81.0 & 73.3 & 73.7 \\
&Names      & 79.4 & 74.0 & \underline{45.5} \\
&Members    & 90.9 & 86.4 & 87.0 \\
&Operators  & 92.9 & 87.7 & 87.7 \\
&Discourse  & 86.2 & 79.6 & \underline{75.3} \\
\bottomrule
\end{tabular}
\caption{F-scores of fine-grained evaluation on the test set of semantic parsers. The evaluation metrics in the table are all based on the Penman format DRG with coarse-grained WordNet synsets.}
\label{table:fine_results}
\end{table}

\textbf{\texttt{Names:}} 
From the results, we observe that the metric \texttt{Names} gives completely opposite results at different evaluation levels. On the node-level, the \texttt{Names} metric in ZH parser scores the lowest, but on the edge-level, \texttt{Names} metric in ZH$\to$EN$_{zh}$ gives the lowest scores. This is reasonable and expected because the node-level \texttt{Names} metric only evaluates whether the parser can parse concepts to contain named entities, so the results of ZH$\to$EN$_{zh}$ parser should be similar to those of the English parser. However, the edge-level \texttt{Names} metric evaluates whether the generated named entities completely match the original text, and the ZH$\to$EN$_{zh}$ parser completely loses the Chinese named entity information.

\textbf{\texttt{Discourse:}} 
An important observation is that the metric \texttt{Discourse} has very low F1 scores on both the node-level and the edge-level for the Chinese parser. Using machine translation and an English parser to parse Chinese (ZH$\to$EN$_{zh}$) will further degrade the performance of the metric \texttt{Discourse}. Based on the text data and parsed output, we find that  discourse relations in Chinese are inconspicuous, and even disappears after being translated into English (see Table~\ref{table:trans} for examples).

\begin{table*}[h]
\centering
\fontsize{10}{12}\selectfont
\resizebox{\textwidth}{!}{
\begin{tabular}{clr}
\toprule
\textbf{Information Type}   &\multicolumn{1}{c}{\textbf{Example}} & \textbf{Lost/Changed in Translation}  \\ 
\midrule
\multirow{6}*{\texttt{Discourse}} & EN: A parrot \textcolor{green}{can} mimic a person's voice. & \multirow{3}*{POSSIBILITY Lost}\\
& ZH: \begin{tabular}[c]{@{}l@{}}\begin{CJK}{UTF8}{gbsn}鹦鹉会模仿人的声音。\end{CJK} \end{tabular} \\
& ZH$\to$EN: Parrots mimic human voices. \\
\cline{2-3}
& EN: Tom asks his mother \textcolor{green}{if} she can buy him a new toy. & \multirow{3}*{ATTRIBUTION Lost}\\
& ZH: \begin{tabular}[c]{@{}l@{}}\begin{CJK}{UTF8}{gbsn}汤姆请求他母亲给他买新玩具。\end{CJK} \end{tabular} \\
& ZH$\to$EN: Tom begged his mother to buy him new toys. \\
\hline
\multirow{9}*{\texttt{Concepts}} 
& EN: That guy is \textcolor{green}{completely} nuts! \ \ ZH: \begin{tabular}[c]{@{}l@{}}\begin{CJK}{UTF8}{gbsn}那家伙真是疯了！ \end{CJK} \end{tabular} &\multirow{2}*{Adverb Lost} \\
& ZH$\to$EN: That guy is crazy!\\
\cline{2-3}
 & EN: She's very \textcolor{green}{handy} with a saw. \ \ ZH: \begin{tabular}[c]{@{}l@{}}\begin{CJK}{UTF8}{gbsn} 她很会用锯子。\end{CJK} \end{tabular}  & \multirow{2}*{Adjective Changed}\\
& ZH$\to$EN: She is \textcolor{red}{good} with a saw. \\
\cline{2-3}
 & EN: I'm \textcolor{green}{awake}. \ \  ZH: \begin{tabular}[c]{@{}l@{}}\begin{CJK}{UTF8}{gbsn} 我醒了。\end{CJK} \end{tabular}  & \multirow{2}*{Adjective Lost}\\
& ZH$\to$EN:  I \textcolor{red}{woke up}. \\
\cline{2-3}
 & EN: Tom is \textcolor{green}{suffering} from a bad headache.  & \multirow{3}*{Verb Changed}\\
 & ZH: \begin{tabular}[c]{@{}l@{}}\begin{CJK}{UTF8}{gbsn} 汤姆头痛得厉害。\end{CJK} \end{tabular}  \\
& ZH$\to$EN:  Tom \textcolor{red}{has} a bad headache. \\
\hline
\multirow{2}*{\texttt{Operators}} & EN: I \textcolor{green}{slept} on the boat. \ \ ZH: \begin{tabular}[c]{@{}l@{}}\begin{CJK}{UTF8}{gbsn} 我睡在船上。\end{CJK} \end{tabular} &\multirow{2}*{Tense Lost}\\
& ZH$\to$EN: I  \textcolor{red}{sleep} on the boat. \\
\hline
\multirow{4}*{\texttt{Negation}} & EN: The music lured \textcolor{green}{everyone}. \ \  ZH: \begin{tabular}[c]{@{}l@{}}\begin{CJK}{UTF8}{gbsn} 音乐吸引了所有人。 \end{CJK} \end{tabular}& \multirow{4}*{NEGATION Lost}\\
& ZH$\to$EN: Music appeals to \textcolor{red}{all}. \\
\cline{2-2}
& EN: The printer \textcolor{green}{doesn't} work. \ \  ZH: \begin{tabular}[c]{@{}l@{}}\begin{CJK}{UTF8}{gbsn} 打印机坏了。 \end{CJK} \end{tabular}  \\
& ZH$\to$EN:The printer is broken. \\
\bottomrule
\end{tabular}}
\caption{Examples of translated English texts with loss of information.}
\label{table:trans}
\end{table*} 

\textbf{\texttt{Concepts:}}
Table~\ref{table:fine_results} shows 
the \texttt{Concepts} scores of ZH parser are lower than those for ZH$\to$EN$_{zh}$ except for the \textit{adj} category.
This is an interesting finding, because the performance of other parts of speech in the ZH parser is worse than that of ZH$\to$EN$_{zh}$, while \textit{adj} is special. We observe that the expressions of adjectives in Chinese translated into English are diverse and may not match the original English text (see Table~\ref{table:trans} and Appendix~\ref{app:output_drs} for relevant examples).

For the English parser, \textit{verbs} are the most difficult words to parse, scoring significantly lower than other parts of speech. However, the difficulty of Chinese semantic parsing is mainly reflected in \textit{adv}.
In addition, the accuracy of ZH$\to$EN$_{zh}$ in parsing concepts of \textit{adv} is significantly better than that of the ZH parser, but it is still the lowest results in four types of parts of speech for ZH$\to$EN$_{zh}$.
On the one hand, the corpus containing adverb data is smaller, which makes the training insufficient. On the other hand, the adverbs in Chinese are usually not obvious and diverse.

For \textit{noun} and \textit{verb}, ZH has the worst performance, with the ZH$\to$EN$_{zh}$ method, the performance of \textit{noun} and \textit{verb} is slightly improved, but it is much worse than the \textbf{EN} parser.
A typical reason is that the English text translated from Chinese may not be consistent with the original English text. We observe that the DRS sequences parsed using the translated text are overall shorter than those parsed using the original English text, some noun concepts are missing, and the verb concepts may be inconsistent with the reference DRS (see Appendix~\ref{app:output_drs} for examples).

\textbf{\texttt{Operators \& Negation:}} 
Our fine-grained results obtained by using machine translation and the English parser are not always worse than training a Chinese parser alone. 
For the metrics \texttt{Negation} and \texttt{Operators}, both methods have similar scores at both the node-level and the edge-level.
However, when we compare the results of ZH$\to$EN$_{zh}$ with EN parser, we find that all the results of ZH$\to$EN$_{zh}$ are significantly lower than those of the EN parser.
We found that tense information is usually lost in the process of English-Chinese translation, but almost no tense information is lost in the process of Chinese-English translation.
This explains why the result of the Chinese parser operator is significantly lower than that of the English parser, while the result of ZH$\to$EN$_{zh}$ is the same as that of the ZH parser. 
For \texttt{Negation}, we can observe something interesting. As the connector NEGATION in English DRss can also express universal quantification (using nesting of two negation operators) for words such as "every" and "always", this information is missing in the translation process, and as a result not picked up by the parser.

\textbf{\texttt{Members \& Roles:}} 
For this metric, ZH$\to$EN$_{zh}$  even slightly outperforms the ZH parser, but they are both lower than the EN parser.
On the one hand, a free translation may lead to a different ordering of semantic information. Although texts with the same meaning but realised with different word order have the same semantic graph, 
a parser based on sequence-to-sequence neural networks may get the wrong graph structure leading to a lower evaluation score of the \texttt{Roles} evaluation metric.
On the other hand, both evaluation metrics are affected by the correctness of \texttt{Concepts}, and in our results, the Chinese parser scored lower than the other two parsers for \texttt{Concepts}.

\section{Conclusion}

Given an annotated meaning bank primarily designed for English, it is feasible to develop a semantic parser for Chinese by pairing the "English" meaning representation with Chinese translations, reaching good results. Most difficulties in Chinese parsing are caused by adverbs, while the diversity of Chinese verbs and adjectives also has a big impact on parsing performance. Using Machine Translation as an alternative to approach semantic parsing for Chinese yields slightly lower results.
Our fine-grained graph evaluation gives better insight when comparing different parsing approaches.

\section*{Acknowledgments}

This work was funded by the NWO-VICI grant ``Lost in Translation---Found in Meaning'' (288-89-003) and the China Scholarship Council (CSC). We thank the anonymous reviewers for detailed comments that improved this paper. We would also like to thank the Center for Information Technology of the University of Groningen for their support and for providing access to the Peregrine high performance computing cluster.

\bibliography{anthology,custom}
\bibliographystyle{acl_natbib}

\appendix
\onecolumn
\section{Result Plots}
According to the fine-grained evaluation results, for both English and Chinese DRS parsing, relatively low f1 scores tend to appear in \texttt{Names} and \texttt{Concepts}. 
The performance of parser declined by approximately five percent after the named entity was converted to Chinese, especially the \texttt{adj} and \texttt{adv}, comparing \textbf{EN} with \textbf{ZH}. 

\begin{figure}[h]
\centering
\includegraphics[scale=.47]{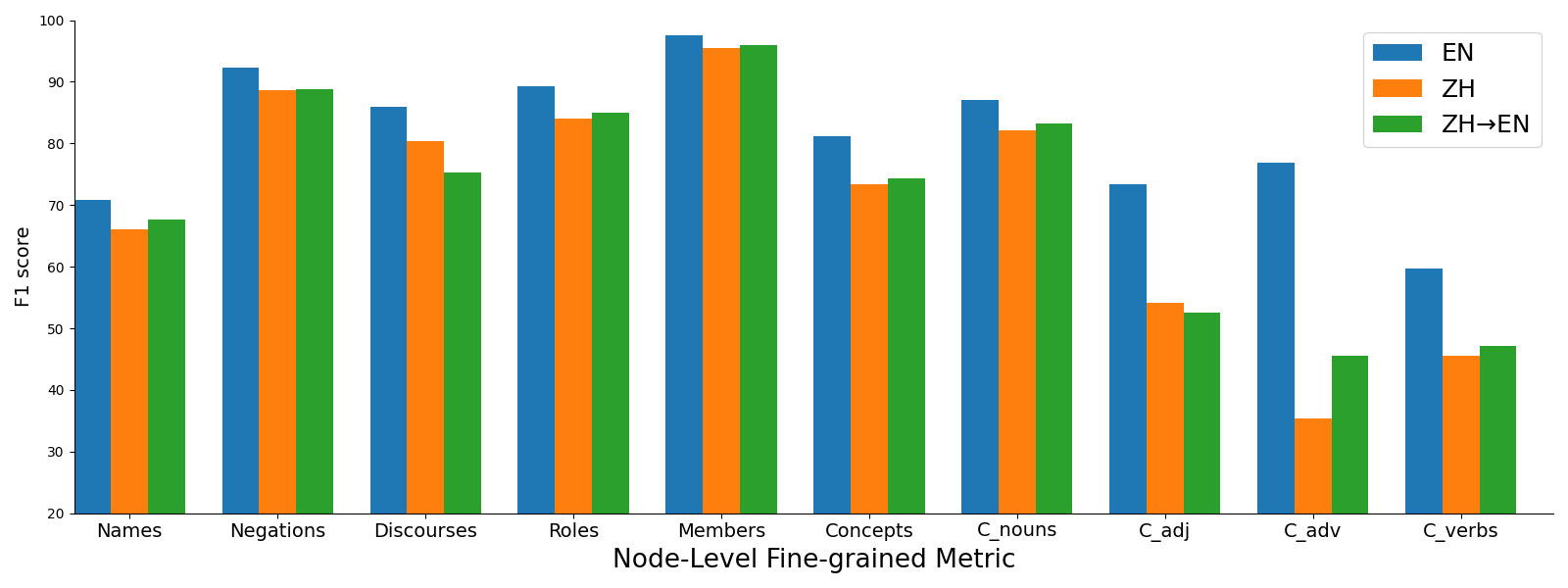}
\caption{Fine-grained results among \textbf{EN}, \textbf{ZH}, and \textbf{ZH$\to$EN$_{zh}$} in Node-level.} 
\label{fig:my_label}
\end{figure}


\section{Output DRS}
\label{app:output_drs}

\begin{table*}[h]
\centering
\resizebox{\textwidth}{!}{
\begin{tabular}{ccp{14cm}}
\toprule
\textbf{Number} & \textbf{Type}  &\multicolumn{1}{c}{\textbf{Example}} \\ 
\midrule
\multirow{10}*{No.1} 
&EN Text & The music \textbf{lured everyone}. \\
&ZH Text &\begin{tabular}[c]{@{}l@{}}\begin{CJK}{UTF8}{gbsn} 音乐吸引了所有人。 \end{CJK} \end{tabular} \\
&ZH$\to$EN &  Music appeals to \textcolor{red}{all}. \\
\cline{2-3}
&\textbf{EN} & music.n.01 \textcolor{blue}{NEGATION} <1 person.n.01 \textcolor{blue}{NEGATION} <1 \textcolor{red}{surprise.v.02} Stimulus -2 Experiencer -1 Time +1 \textcolor{green}{time.n.08 EQU now} \\
&\textbf{ZH} & music.n.01 \textcolor{blue}{NEGATION <1} person.n.01 \textcolor{blue}{NEGATION} <1 \textcolor{red}{appeal.v.01} Agent -2 Theme -1 Time +1 \textcolor{green}{time.n.08 TPR now} \\
&\textbf{ZH$\to$EN$_{zh}$} & event.v.01 Participant +1 music.n.01 \textcolor{red}{appeal.v.01} Theme -1\\
\cline{2-3}
&\textbf{Gold DRS} & music.n.01 NEGATION <1 person.n.01 NEGATION <1 lure.v.01 Agent -2 Patient -1 Time +1 time.n.08 TPR now \\
\midrule
\midrule
\multirow{10}*{No.2} 
&EN Text & She's \textbf{very handy} with a saw. \\
&ZH Text & \begin{tabular}[c]{@{}l@{}}\begin{CJK}{UTF8}{gbsn} 她很会用锯子。\end{CJK} \end{tabular} \\
&ZH$\to$EN &She is \textcolor{red}{good} with a saw. \\
\cline{2-3}
&\textbf{EN} & female.n.02 time.n.08 EQU now \textcolor{blue}{very.r.01} \textcolor{green}{handy.a.01} AttributeOf -3 Time -2 Degree -1 Instrument +1 saw.n.02\\
&\textbf{ZH} & female.n.02 time.n.08 TSU now \textcolor{red}{use.v.01} Agent -2 Time -1 Theme +1 Instrument +2 entity.n.01 saw.n.02\\
&\textbf{ZH$\to$EN$_{zh}$} & female.n.02 time.n.08 EQU now \textcolor{red}{good.a.01} AttributeOf -2 Time -1 Instrument +1 saw.n.02\\
\cline{2-3}
&\textbf{Gold DRS} &  female.n.02 time.n.08 EQU now very.r.01 handy.a.03 AttributeOf -3 Time -2 Degree -1 Instrument +1 saw.n.02\\
\bottomrule
\end{tabular}}
\caption{Examples of output DRSs by different parsers.}
\label{table:output_drs}
\end{table*}

\end{document}